\documentclass[letterpaper, 10 pt, conference]{ieeeconf}  

\IEEEoverridecommandlockouts                              

\usepackage{graphicx} 
\usepackage{amsmath} 
\usepackage{amssymb}  
\usepackage{bm}
\usepackage{booktabs}
\usepackage{float}
\usepackage{flushend}
\usepackage{pifont}
\usepackage{tabularx}
\usepackage[backend=biber,style=numeric,sorting=none]{biblatex}
\usepackage[bookmarks=true]{hyperref}

\newcommand{\xmark}{\ding{55}}

\newcommand{\real}{\mathbb{R}}

\title{\LARGE \bf
NeuralLabeling: A versatile toolset for labeling vision datasets using Neural Radiance Fields
}

\author{Floris Erich$^{1*}$, Naoya Chiba$^{2}$, Abdullah Mustafa$^{1}$, Yusuke Yoshiyasu$^{1}$, \\Noriaki Ando$^{1}$, Ryo Hanai$^{1}$, Yukiyasu Domae$^{1}$
\thanks{* Corresponding author, reachable at firstname.lastname@aist.go.jp.}%
\thanks{$^{1}$National Institute of Advanced Industrial Science and Technology, Tokyo, Japan.}%
\thanks{$^{2}$Tohoku University, Sendai, Japan.}%
}

\begin{document}

\maketitle
\thispagestyle{empty}
\pagestyle{empty}

\begin{abstract}
  We present \emph{NeuralLabeling}, a labeling approach and toolset for annotating 3D scenes using either bounding boxes or meshes and generating segmentation masks, affordance maps, 2D bounding boxes, 3D bounding boxes, 6DOF object poses, depth maps, and object meshes.
  \emph{NeuralLabeling} uses Neural Radiance Fields (NeRF) as a renderer, allowing labeling to be performed using 3D spatial tools while incorporating geometric clues such as occlusions, relying only on images captured from multiple viewpoints as input.
  To demonstrate the applicability of \emph{NeuralLabeling} to a practical problem in robotics, we added ground truth depth maps to 30000 frames of transparent object RGB and noisy depth maps of glasses placed in a dishwasher captured using an RGBD sensor, yielding the Dishwasher30k dataset.
  We show that training a simple deep neural network with supervision using the annotated depth maps yields a higher reconstruction performance than training with the previously applied weakly supervised approach.
  We also show how instance segmentation and depth completion datasets generated using NeuralLabeling can be incorporated into a robot application for grasping transparent objects placed in a dishwasher with an accuracy of 83.3\%, compared to 16.3\% without depth completion.
  Supplementary URI: \href{https://florise.github.io/neural_labeling_web/}{https://florise.github.io/neural\_labeling\_web/}.
\end{abstract}

\section{INTRODUCTION}

Deep learning requires large datasets, which are time-intensive and expensive to create.
There are various approaches to avoid this, such as using foundation models or weakly supervised training methods like cyclic adversarial learning~\cite{Zhu_2017_ICCV}.
However, despite being trained on massive datasets, foundation models such as Segment Anything~\cite{kirillov2023} and CLIP~\cite{radford2021a} still rely on inference data to be similar to the training data, which is not always the case.
Models trained using weakly supervised learning might outperform state-of-the-art models when the SOTA models are not trained on task-specific data, but their performance is lower than SOTA models evaluated on evaluation data more similar to their training data.
Thus there is a need for tools that can support large dataset creation in a time-efficient low-cost manner.
We hope to contribute to solving this problem by introducing a labeling tool for computer vision datasets that uses the power of Neural Radiance Fields (NeRF)~\cite{mildenhall2020} for photorealistic rendering and geometric understanding.
Because 3D Vision can take advantage of 3D consistency, labels on a single scene can be applied to images from multiple viewpoints.
This property works particularly well with photorealistic renderings such as NeRF, where richly annotated data with many views is available with only simple manual 3D labeling.
This not only saves significant labeling time but is also useful in automatically generating a consistent dataset.

Specialized labeling tools are essential for labeling vision datasets, and both academic researchers and commercial entities have released such tools.
Most existing labeling tools (such as Segment Anything Labeling Tool~\cite{salt} and Roboflow~\cite{roboflow}) use single images and therefore require significant human effort to annotate long sequences, use sequential data but have no geometric understanding so they cannot be used for annotating 6DOF poses~\cite{cheng2022xmem}, or require depth data to obtain geometric information~\cite{lai2012,zimmer2019,singh2021}.
Our toolkit, \emph{NeuralLabeling}, operates on sequences of images and can thus be used to more rapidly label large datasets.
By using manual scaling and NeRF depth reconstruction~\cite{mildenhall2020}, NeuralLabeling does not rely on input depth data except when used for generating datasets for depth completion tasks.
Due to improvements in the training time of NeRFs~\cite{muller2022}, \emph{NeuralLabeling} does not rely on slow dense mesh reconstruction and instead only requires camera pose estimation, which takes around an hour per scene of approximately 500 images, and could be further reduced by selecting key frames and interpolating camera poses between them~\cite{takeda2023} or avoided using NeRF recording applications such as NeRFCapture~\cite{NeRFCapture}.

\begin{figure}
  \centering
  \includegraphics[scale=0.58]{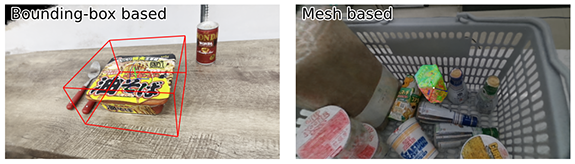}
  \caption{\emph{NeuralLabeling} supports two pipelines for labeling NeRFs: Bounding-box-based labeling for uncluttered scenes and mesh-based labeling for cluttered scenes.}
  \label{fig:labeling}
\end{figure}

This paper has two main contributions:
(1) We present \emph{NeuralLabeling}, a novel labeling system that is deeply integrated into a NeRF-based photorealistic rendering system (Section~\ref{ref:methodology}).
(2) We construct the Dishwasher30k dataset, which can be used for NeRF-based transparent object depth completion research, and release it on our web page. 
Furthermore, we perform the following experiments to validate our approach:
(1) We evaluate the accuracy of \emph{NeuralLabeling} for generating transparent object datasets for depth completion (Section~\ref{sec:groundtruth}).
(2) We evaluate the accuracy of \emph{NeuralLabeling} for generating datasets for object segmentation, taking into account occlusions (Section~\ref{sec:nerfocclusion}).
(3) We demonstrate how training a transparent object depth completion network using a dataset generated by \emph{NeuralLabeling} leads to improved performance compared to unsupervised datasets (Section~\ref{sec:depthcompletionexperiment}).
(4) We show that networks trained using datasets generated by \emph{NeuralLabeling} can be integrated into a robot manipulation system (Section~\ref{sec:robotexperiment}).

\begin{figure*}
  \vspace{0.5em}
  \centering
  \includegraphics[scale=1]{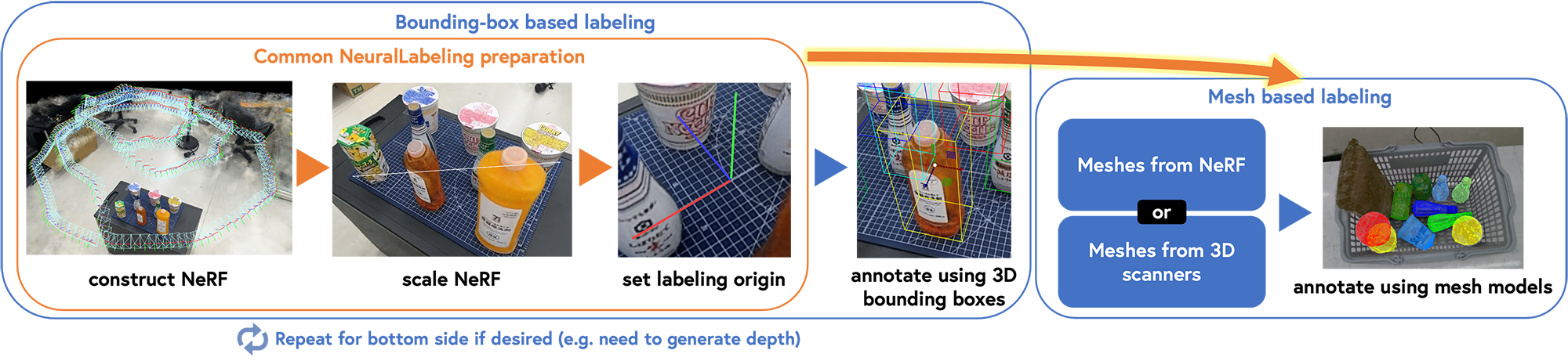}
  \caption{A scene can be labeled using either bounding-boxes or using meshes. Bounding boxes can be used to extract meshes from a scene.}
  \label{fig:workflow}
\end{figure*}

\begin{table}
    \vspace{0.4em}
  \caption{Comparing unique aspects of labeling tools. All tools support segmentation masks. NDR~=~No Input Depth Required, G~=~Geometry, M~=~Mesh, 6D~=~6DOF poses, O~=~Occlusion masks, A~=~Affordance maps, OD~=~Object Depth} 
  \centering
\begin{tabular}{l | c | c c | c c c c }
\toprule
                                  & Inputs     & \multicolumn{2}{c|}{Selection} & \multicolumn{4}{c}{Outputs} \\
Tool                              & NDR         & G          & M                 & 6D & O & A      & OD \\
\midrule
ProgressLabeller~\cite{chen2022d} & \checkmark & \xmark     & \checkmark        & \checkmark & \xmark     & \xmark     & \xmark \\
3D-DAT~\cite{suchi2023}           & \checkmark & \xmark     & \checkmark        & \checkmark & \xmark     & \xmark     & \xmark \\
Nerfing It~\cite{blomqvist2023b}  & \checkmark & \checkmark & \xmark            & \xmark     & \xmark     & \xmark     & \xmark \\
RapidPoseLabels~\cite{singh2021}  & \xmark     & \checkmark & \checkmark        & \checkmark & \xmark     & \xmark     & \xmark \\
HANDAL~\cite{guo2023}             & \xmark     & \checkmark & \checkmark        & \checkmark & \checkmark & \checkmark & \xmark \\
NeuralLabeling (Ours)             & \checkmark & \checkmark & \checkmark        & \checkmark & \checkmark & \checkmark & \checkmark \\
\bottomrule
\end{tabular}
\label{tab:labeling_tools}
\end{table}

\section{BACKGROUND}

\subsection{Vision data labeling tools}

\begin{figure*}
  \centering
  \includegraphics[scale=0.5]{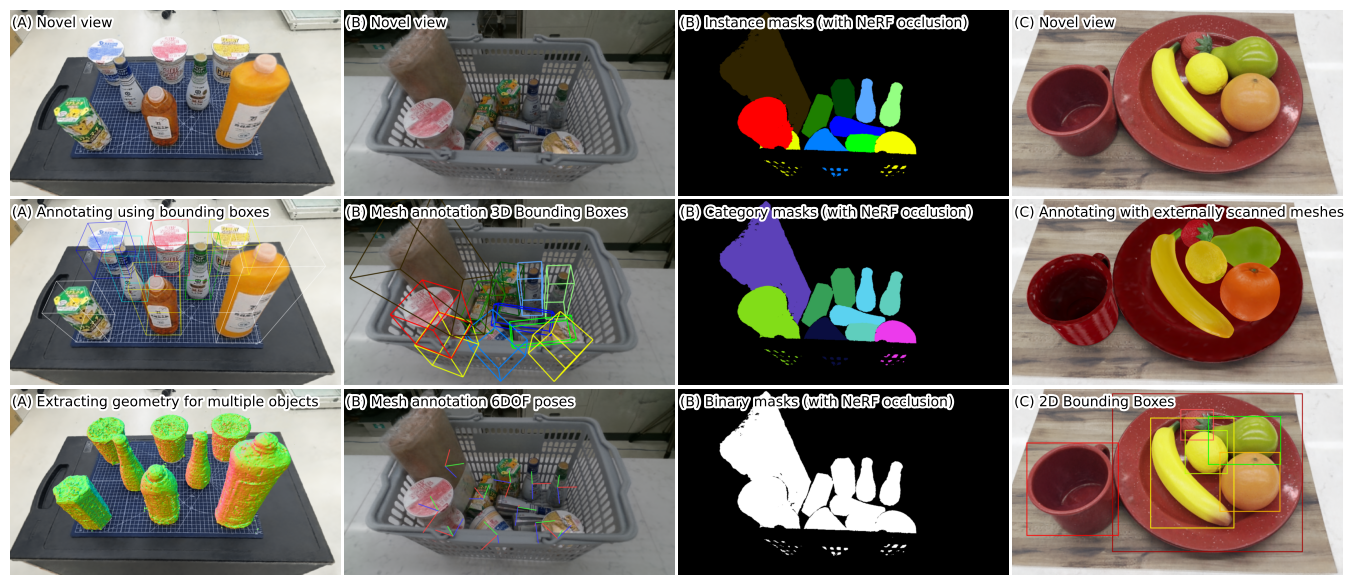}
  \caption{
  \emph{NeuralLabeling} supports a wide variety of outputs.
  Circled letter references the scene: (A) Mostly Lambertian objects placed upright for mesh extraction, second row shows the annotated bounding boxes, third row shows the geometry generated using the bounding boxes.
  (B) Most of the objects from (A) placed in a shopping basket and annotated using the meshes generated from (A), towel was captured separately, second row shows 3D bounding boxes based on the mesh annotations, third row shows 6DOF poses based on the mesh annotations. Second column of (B) shows instance masks, category masks and binary masks, each using NeRF-to-mesh occlusions rendered directly by \emph{NeuralLabeling} to improve segmentation accuracy.
  (C) Lambertian objects placed on a lunch plate. We use YCB objects for which we use openly available meshes based on 3D scans using the Google Scanner, second row shows the meshes rendered directly in the scene, third row shows 2D bounding boxes generated based on mesh geometry.}
  \label{fig:outputs}
\end{figure*}

\emph{NeuralLabeling} was inspired by various recent tools for creating labeled datasets but qualitatively improves upon each of them.
ProgressLabeller~\cite{chen2022d} is a state-of-the-art labeling tool that uses mesh alignment and posed camera images.
RapidPoseLabels~\cite{singh2021} is an RGBD-based labeling tool, allowing for labeling objects with pose annotations.
Because it uses RGBD data as input it cannot be used if depth data is unavailable or unreliable.
3D-DAT~\cite{suchi2023} is a mesh-based labeling tool implemented as a Blender plugin.
It uses NeRF for automated alignment of objects with NeRF geometry, but it requires meshes to be provided as input.
It also does not support NeRF-to-mesh occlusions.
Nerfing It~\cite{blomqvist2023b} is a NeRF-based labeling tool, but it does not support mesh-based labeling.
It also uses a vanilla NeRF implementation that is not optimized for speed, and thus requires long training times to prepare scenes for labeling.
Table~\ref{tab:labeling_tools} compares \emph{NeuralLabeling} with various state-of-the-art labeling tools.

Our work resembles the pipeline used for preparing the HANDAL dataset~\cite{guo2023}.
Their work uses a bi-methodical 3D-bounding-box-based and mesh-based labeling approach, similar to what we present in this paper.
An advantage of HANDAL is that it also supports labeling dynamic scenes.
An advantage of our tool is that it can be used to generate depth maps for transparent objects.
NeuralLabeling can generate segmentation masks that can be used for training neural networks to perform object segmentation, whereas the HANDAL pipeline relies on segmentation masks generated using a pre-trained tracker~\cite{cheng2022xmem}.
Their work uses automatic scaling based on depth input, whereas our work relies on manual scaling using a scaling tool.
Inspired by their work, we added an affordance labeling tool to \emph{NeuralLabeling}.

NeuralLabeling enables the labeling of existing scenes using NeRF, however in the parallel work PEGASUS~\cite{meyer2024pegasus} we allow generating datasets by inserting objects into an existing scene and rendering them using 3D Gaussian Splatting.
By inserting custom objects into a scene, a wider variety of object configurations can be generated, thus leading to more variety in the generated datasets.
However, the PEGASUS renderer is unaware of scene-specific lighting, whereas for NeuralLabeling the objects inherit natural scene lighting.

\subsection{Transparent Object Depth Completion}
\emph{NeuralLabeling} started as a tool to label transparent objects with accurate depth estimates to enable robots to estimate depth and shape of glasses and cups, without relying on expensive photorealistic simulations.
Deep learning approaches have greatly contributed to solving the problem of transparent object depth completion~\cite{sajjan2020clear}, however most existing datasets consist of glasses placed in simple environments such as on tables and floors~\cite{sajjan2020clear,chen2022clearpose,zhu2021a}.
State-of-the-art pretrained models underperform when applied to more complex environments such as a dishwasher~\cite{erich2023fakingdepth}.
Weakly supervised training methods can outperform state-of-the-art supervised models, but still underperform compared to the performance of the state-of-the-art models on data that is more similar to their training data.
We show that \emph{NeuralLabeling} can be used to easily create supervised datasets for a complex environment such as a dishwasher, and that a network trained on such a supervised dataset can outperform a network trained on a weakly supervised dataset.
Using \emph{NeuralLabeling}, it took roughly one workweek to construct this dataset, which contains NeRFs, mesh models, alignment configurations of the meshes with the NeRF, generated depth, and generated segmentation masks.
We release this dataset, which we name \emph{Dishwasher30k}.

\section{METHODOLOGY}
\label{ref:methodology}

We support labeling using either 3D bounding-boxes or meshes (Fig.~\ref{fig:labeling}).
3D-bounding-box-based labeling is useful when scenes are uncluttered and/or high quality object meshes for applying labels to the scene are not available.
Mesh-based labeling is useful when scenes are cluttered or if we already have object meshes available.
We support mesh extraction using bounding-boxes, which enables a novel pipeline where we obtain mesh models for objects placed in an uncluttered manner, and then reuse these models in a cluttered scene.
Fig.~\ref{fig:workflow} gives a more detailed overview of the combined labeling pipeline.

We aim to generate semantic segmentation masks, 2D and 3D bounding boxes, 6DOF object poses, depth maps and object meshes for each frame in a RGB image sequence (Fig.~\ref{fig:outputs}).
Segmentation masks are further classified into binary, instance and class segmentation masks.
2D bounding boxes are defined by the lower left corner and upper right corner.
3D bounding boxes are defined by the lower left front corner, upper right back corner and an orientation.
When using the 3D-bounding-box-based labeling workflow, we can either directly use the labeled bounding-boxes or we can optimize the bounding-boxes to tightly fit their geometry.
6DOF object poses are defined by translation and rotation of objects relative to the camera pose.
Depth maps are defined by depth elements of rays cast perpendicular from the camera plane to the nearest surface, or $0$ if no nearest surface exists for a depth element.
Object meshes are defined using the common Wavefront OBJ format~\cite{zotero-3323}.
In the downstream tasks presented in this paper we use object meshes, semantic segmentation masks and depth maps.
The other output types were added to increase the flexibility of the toolset.
To annotate an object with affordances, sub-bounding-boxes can be added, which are stored as a JSON file alongside the exported geometry and automatically loaded when inserting exported meshes in new scenes.
Per-object affordance maps can be exported in a similar way as segmentation masks.

\subsection{Uncluttered scene pipeline}
In this pipeline an uncluttered scene is annotated using bounding-boxes.

\begin{enumerate}
  \item Record RGB frames of a scene containing objects to label: $\mathbf I \in \real^{N \times W \times H \times 3}$, where $N$ is number of frames, $W$ is width and $H$ is height.
  \item Obtain camera extrinsics $\mathbf T \in \real^{3 \times 4} = [ \mathbf R\, |\, \vec t]$ and intrinsics for each frame using Structure-from-Motion algorithms such as COLMAP~\cite{schoenberger2016sfm,schoenberger2016mvs} or hloc~\cite{sarlin2019coarse}, where $\mathbf R$ is camera rotation matrix and $\vec t$ is camera translation vector.
  \item Determine scale $s$ by comparing keypoints or using AR marker~\cite{meyer2023}, and rescale positions $\mathbf T_s = [\mathbf R\, |\, s \cdot \vec t]$.
  \item Render NeRF using $\mathbf T_s$ and $\mathbf I$.
  \item Label objects using bounding-boxes, by inserting boxes, translating and rotating them to surround target objects.
  \item Export geometry contained in bounding-boxes by querying density of NeRF in bounding-box areas, apply density filter and run marching cubes~\cite{lorensen1987}.
\end{enumerate}

\subsection{Cluttered scene pipeline}
In this pipeline, a cluttered scene is labeled using polygonal meshes.
If we have access to the physical objects in the scene, meshes can be obtained through the uncluttered scene pipeline.
This pipeline repeats steps 1-4 from the uncluttered scene pipeline but replaces steps 5 and 6 with the following:

\begin{enumerate}
  \setcounter{enumi}{4}
  \item Label objects using mesh models, by inserting meshes, translating and rotating them to be aligned with NeRF rendering of objects.
  \item Export semantic segmentation masks, 2D and 3D bounding boxes, 6DOF object poses and depth maps.
\end{enumerate}

\subsection{Implementation details}
Because our labeling functionality is specialized, we implemented \emph{NeuralLabeling} as a fork of \emph{instant-ngp}~\cite{muller2022} instead of merging our changes into the main project.
\emph{instant-ngp} allows for parallel training and rendering, and with our fork also for labeling.
Rendering of geometry extracted using marching cubes and rendering of (re)inserted meshes is implemented using OpenGL.
In the bounding-box-based pipeline, we support real-time geometry previews from NeRF.
Rendering of overlay effects such as 2D and 3D bounding boxes is implemented using ImGui\footnote{Online: \url{https://github.com/ocornut/imgui}}.
Manipulating objects (translation, rotation, scaling) is implemented using ImGuizmo\footnote{Online: \url{https://github.com/CedricGuillemet/ImGuizmo}}.
We implemented an accelerated algorithm for object alignment, using multi-threading and CUDA kernels.
NeRF-to-mesh occlusions are handled by comparing estimated depth of rays traced in the NeRF rendering with depth of the rendered meshes fragments.
We integrate improved transparent object depth estimation via Dex-NeRF~\cite{ichnowski2021a}.
We enable scripting of \emph{NeuralLabeling} through Python bindings, which is useful for automated dataset generation.

\section{EVALUATION}

Our toolkit can be used to easily and quickly label photorealistic scenes that would be hard to manually model and generate various useful outputs for downstream deep learning tasks.
We demonstrate this by (1) evaluating base performance of depth generation using object annotations in Section~\ref{sec:groundtruth}, (2) evaluating segmentation performance of annotating opaque objects with a high degree of environment occlusion in Section~\ref{sec:nerfocclusion}, (3) annotating scenes containing transparent objects in a complex environment and training neural networks using generated data in Section~\ref{sec:depthcompletionexperiment} and (4) evaluating the performance of a holistic robotic transparent object manipulation system using neural networks trained or fine-tuned on generated datasets in Section~\ref{sec:robotexperiment}.

\subsection{Ground truth depth label accuracy}
\label{sec:groundtruth}

\begin{figure}
  \includegraphics[scale=0.62]{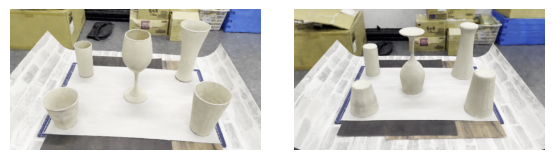}
  \caption{Opaque clones of glasses placed up- and down-facing, rendered using NeRF. Using the bounding-box labeling pipeline we extract meshes that are used for annotating the dishwasher scenes.}
  \label{fig:glasses}
\end{figure}

To evaluate the optimal performance of using NeuralLabeling, we recorded 30 samples of color and depth data collected from glasses placed into a scene.
The glass is then replaced with an opaque clone placed into the same position and the depth data is recaptured.
To place the opaque clone into the same position as the original glass, we take a picture of the original scene using a camera and render an overlay image.
This is a typical approach for creating real-world validation data for transparent object depth completion~\cite{sajjan2020clear}.
In addition to capturing ground truth depth by manually aligning opaque clones, we also captured a NeRF scene recording.
We create meshes of the glasses by recording two environments with opaque clones of the glasses placed facing upwards and downwards (Fig.~\ref{fig:glasses}).
We used opaque clones instead of the original glasses for creating meshes, as this produced higher quality meshes due to NeRF not being able to correctly estimate the inner surface of the original glasses.
We applied our cluttered scene pipeline to this dataset to generate experimental data by aligning the scanned meshes with the original glasses.
To evaluate the accuracy of the pipeline, we compare the generated depth of the labels applied to the transparent scene with the ground truth depth generated from aligning opaque clones.
This experiment resulted in a median error of 4mm and an MAE (mean absolute error) of 9mm at a mean working distance of 649mm (1.4\% relative error), which is similar to the stated depth estimation error of the depth sensor (2\% at 2 meters working distance).
Exporting depth using NeuralLabeling without mesh annotations, but using Dex-NeRF-like~\cite{ichnowski2021a} transparent object depth estimation resulted in a median error of 5mm and an MAE of 16mm (2.4\% relative error).
We can conclude that our method for labeling transparent object depth is at least as accurate as the applied depth sensor is on opaque objects, and more accurate compared to Dex-NeRF.

\begin{table*}
    \vspace{0.4em}
    \centering
    \caption{Quantitative results of masking using NeRF occlusion. Higher score is better.}
    \begin{tabular}{c | c c c c c | c c c c c}
    \toprule
           & \multicolumn{5}{c|}{Binary} & \multicolumn{5}{c}{Category} \\
    Method & F1-score & IoU  & Accuracy & Precision & Recall & F1-score & IoU  & Accuracy & Precision & Recall \\
    \midrule
    SAM    & 0.80     & 0.67 & 0.97     & 0.71      & 0.90   & 0.74     & 0.61 & 1.00     & 0.68      & 0.85 \\
    XMem   & 0.85     & 0.74 & 0.98     & 0.82      & 0.88   & 0.80     & 0.68 & 1.00     & 0.77      & 0.84 \\
    Ours   & 0.83     & 0.70 & 0.98     & 0.95      & 0.73   & 0.80     & 0.68 & 1.00     & 0.93      & 0.71 \\
    \bottomrule
    \end{tabular}
    \label{tab:quantitative_results_masking}
\end{table*}

\subsection{NeRF occlusion for generating segmentation masks}
\label{sec:nerfocclusion}
One of the unique functions of \emph{NeuralLabeling} is to use NeRF occlusions to generate accurate segmentation masks.
We performed a small experiment to measure the effectiveness.
We labeled a sequence of three heavily occluded frames of the basket scene (scene B of Fig.~\ref{fig:outputs}) with ground truth segmentation masks, then calculated F1-score, Intersection-over-Union (IoU), accuracy, precision and recall.
We compare our method with Segment Anything (SAM)~\cite{kirillov2023} labels using 2D bounding boxes and XMem~\cite{cheng2022xmem} by using the first frame as input.
Quantitative and qualitative results can be found in Table~\ref{tab:quantitative_results_masking} and in the supplemental materials respectively.
Our method outperforms SAM in almost every metric, while performing similar to XMem.
Some qualitative benefits of our approach are that \emph{NeuralLabeling} does not require all objects to be visible in the first frame of a sequence such as with XMem and does not need per frame 2D bounding boxes such as with SAM.
For generating NeRF occlusions we rely on extracting an accurate depth estimate from NeRF, which is difficult for objects with highlights and reflections.
Our method for example struggles to generate accurate segmentation masks of the towel from scene B, which is wrapped in plastic.
Compared to the other methods, the segmentation masks generated by NeuralLabeling are more conservative, which decreases the recall score.

\subsection{Training networks for depth completion}
\label{sec:depthcompletionexperiment}

\begin{figure*}
  \centering
  \includegraphics[width=\textwidth]{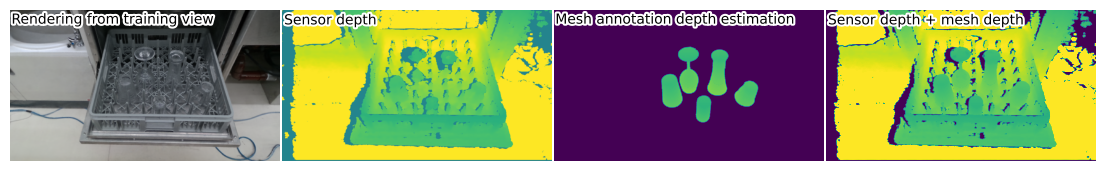}
  \caption{Non-Lambertian objects in a complicated environment, annotated using opaque clone NeRF meshes, second column shows sensor depth estimate using RealSense D415, third column shows estimated object depth based on mesh annotations, fourth column shows the combination of generated depth with noisy sensor depth, which can be used as ground truth data for training a deep neural network.}
  \label{fig:depth_gt}
\end{figure*}

In a previous study~\cite{erich2023fakingdepth} we evaluated the usage of unpaired training data with a cyclic adversarial training approach~\cite{Zhu_2017_ICCV} for transparent object depth completion.
We use the same dataset and network design from the previous study but added supervised ground truth depth maps and instance segmentation masks using \emph{NeuralLabeling}.
The RGB images from the original dataset were used for determining camera poses and NeRF rendering.

\subsubsection{Dataset preparation}
For transparent objects a marching cubes threshold can be used to tune the mesh geometry similar to Dex-NeRF~\cite{ichnowski2021a}, however the observed mesh quality was still lower than using opaque clones.
We merged the upwards and downwards facing meshes using MeshLab~\cite{journals/ercim/CignoniCR08} to produce complete meshes of the glasses.
We want to show that good results can be obtained using low cost methods, so we avoided using more advanced techniques such as using expensive camera setups~\cite{erich2023neuralscanning} or commercial 3D scanners.

The meshes are manually aligned with the NeRF rendering.
We generated camera pose estimates for 59 out of 60 scenes, camera pose estimation failed on one scene.
We calibrated the camera pose scales for each scene by measuring the distance between two points where the real world distance was known, taking about a minute per scene.
It then took two working days to label the 59 scenes with the meshes.
An automated process generated the depth maps for the 59 scenes, which took around three minutes per scene.

NeRF-to-mesh occlusions could not reliably be generated due to the difficulty in estimating the depth elements of inner surfaces of glasses using NeRF.
Instead, we use sensor depth elements to occlude the generated depth elements.
Sensor depth elements for transparent objects are inaccurate due to missing elements ($depth=0$), background depth elements and noisy surface depth elements.
By using sensor depth elements for calculating occlusions, we can fill in missing depth elements and correct background depth elements to be on the object surface, but some noisy surface depth elements that were inaccurately estimated as being too close to the camera might remain.
Fig.~\ref{fig:depth_gt} shows a sample from the dishwasher dataset, with the original depth recorded by the depth sensor, with generated depth estimate from mesh annotations, and finally, the combined sensor and mesh depth that is used as ground truth for training our network.

We reuse the validation set of the original paper~\cite{erich2023fakingdepth} ($N=26$), containing scenes in which glasses were manually aligned with opaque clones (using the same process described in Section~\ref{sec:groundtruth}).
All evaluation samples are patches with dimensions $512 \times 512 \times C$ extracted from the center of the sensor frame with dimensions $1280 \times 720 \times C$, where depth is clipped to the $[450,2000]$ mm range.
For training we first random crop frames horizontally to dimensions $720 \times 720 \times C$ and then resize to dimensions $512 \times 512 \times C$ using nearest neighbor interpolation.
For evaluation we center crop the frames horizontally before resizing.
$C$ is the number of channels for the modality: $1$ for depth-only, $3$ for RGB only, $4$ for RGBD.
For each channel we map the values to the domain $[-1,1]$ from the original domains $[0,255]$ from RGB and $[450,2000]$ for depth.

\subsubsection{Networks and training}
Whereas in the previous study we used two generator networks and two discriminator networks, in this study we use only a single generator network for each evaluated modality.
The generator network in the previous study and the current study is a simple U-Net, based on Pix2pix~\cite{isola2017image}.
We evaluated three modalities: RGBD2Depth, Depth2Depth and RGB2Depth.
In the previous study we evaluated Depth2Depth and RGBD2RGBD modalities (the method required input and output type to be symmetric, so the target output was RGBD data of scenes containing opaque clones of the original glasses by spray painting them).
In both the original study and the current study we trained for $2 \times 10^6$ iterations.

\subsubsection{Results}

\newcommand{\B}[1]{\emph{#1}}

\begin{table*}
    \vspace{0.4em}
  \caption{Transparent object depth completion using weakly supervised methods versus strongly supervised methods} 
  \begin{tabularx}{0.95\textwidth}{X X c c c c c c}
    \toprule
    Training regime          & Modality    & RMSE (m) ↓ & MAE (m) ↓ & Rel ↓     & 1.05 ↑    & 1.10 ↑    & 1.25 ↑ \\
    \midrule
    Joint Bilateral Filter   & RGBD2Depth  & 0.067      & 0.048     & 0.083     & 0.477     & 0.688     & 0.950 \\ 
    ClearGrasp               & RGBD2Depth  & 0.090      & 0.057     & 0.120     & 0.404     & 0.555     & 0.840  \\
    \midrule
    Cyclic adversarial       & RGBD2RGBD   & 0.061      & 0.040     & 0.072     & 0.528     & 0.767     & 0.940 \\
    Cyclic adversarial       & Depth2Depth & 0.058      & 0.035     & 0.061     & 0.589     & 0.861     & 0.954 \\
    \midrule
    Dishwasher30k supervised & RGBD2Depth  & \B{0.037}  & 0.023     & 0.039     & 0.725     & 0.880     & \B{0.959} \\
    Dishwasher30k supervised & Depth2Depth & 0.043      & \B{0.021} & \B{0.038} & \B{0.800} & \B{0.895} & 0.955 \\
    Dishwasher30k supervised & RGB2Depth   & 0.045      & 0.028     & 0.049     & 0.676     & 0.861     & 0.948 \\
    \bottomrule
  \end{tabularx}
\label{tab:tode}
\end{table*}

\newcommand{\makeresrow}[3]{
\begin{minipage}[t]{.14\textwidth}
    \centering
    \includegraphics[width=0.9\textwidth]{results/supervised/#1-input-color.png}
\end{minipage}
&
\begin{minipage}[t]{.14\textwidth}
    \centering
    \includegraphics[width=0.9\textwidth]{results/supervised/#1-input-depth.png}
\end{minipage}
&
\begin{minipage}[t]{.14\textwidth}
    \centering
    \includegraphics[width=0.9\textwidth]{results/supervised/#1-output-depth.png}
    #2
    \vspace{0.2em}
\end{minipage}
&
\begin{minipage}[t]{.14\textwidth}
    \centering
    \includegraphics[width=0.9\textwidth]{results/old_method//#1-depth-output.png}
    #3
    \vspace{0.2em}
\end{minipage}
&
\begin{minipage}[t]{.14\textwidth}
    \centering
    \includegraphics[width=0.9\textwidth]{results/old_method/#1-depth-ground-truth.png}
\end{minipage}
\\
}
 
\begin{table*}
    \vspace{1mm}
    \centering
    \caption{Qualitative results of our supervised method and previous best cyclic adversarial method.}
    \begin{tabular}{c c c c c}
    \toprule
    Captured Color & Captured Depth & Our result & CycleGAN result & Ground truth depth \\
    \midrule
    \multicolumn{5}{c}{Three samples with the lowest MAE using our method}\\
    \midrule
    \makeresrow{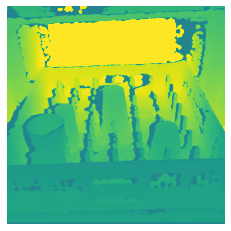}{0.012}{0.032}
    \makeresrow{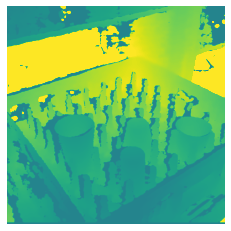}{0.013}{0.025}
    \makeresrow{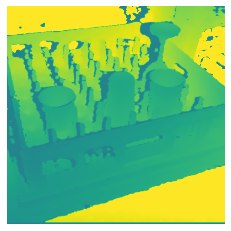}{0.014}{0.033}
    \midrule
    \multicolumn{5}{c}{Three samples with the highest MAE using our method}\\
    \midrule
    \makeresrow{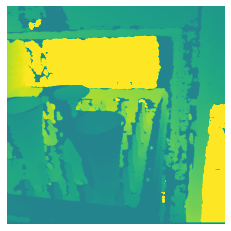}{0.034}{0.066}
    \makeresrow{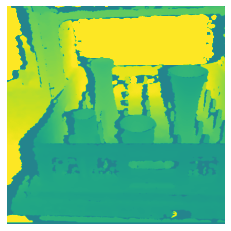}{0.035}{0.054}
    \makeresrow{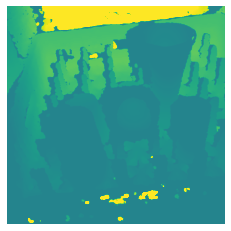}{0.036}{0.050}
    \bottomrule
    \end{tabular}
    \label{tab:qualitative_results}
\end{table*}

Table~\ref{tab:tode} and Table~\ref{tab:qualitative_results} contain quantitative and qualitative results.
Cyclic adversarial measurements are sourced from our previous paper~\cite{erich2023fakingdepth}.
Metrics used are Root Mean Square Error (RMSE), Mean Absolute Error (MAE), Relative error (Rel), proportion of depth elements with less than 5\% error (1.05), less than 10\% error (1.10) and less than 25\% error (1.25).
We apply the metrics to the depth elements covered by transparent objects.
Regardless of the modality used, we could obtain a significant improvement by using supervised data created using \emph{NeuralLabeling}.

The weakly supervised approach required recording a separate dataset containing 60 scenes of opaque clones, which took about 4 hours to collect, but our current method does not use this.
For the current supervised approach, we had to record two scenes of opaque clones to extract meshes, and then label the original transparent objects in the dishwasher scenes.
Creating opaque meshes took around 8 hours.
Aligning the opaque meshes with the transparent scenes took around 16 hours, the time per scene varying based on the amount of objects in the scene.
Training time for the original approach was around 4 times longer, as four networks had to be trained instead of a single network.
NeuralLabeling requires COLMAP camera estimates taking around an hour per scene and we pretrained the NeRFs to allow for faster labeling for around an hour per scene.
Predictions using the newly trained networks are slightly more blurry than the CycleGAN approach due to not using a discriminator, but because the depth maps are for robot consumption this was not considered an issue.
We conclude that the \emph{NeuralLabeling} approach requires more time to prepare the dataset but allows for more accurate depth estimates and efficient training for downstream tasks.

\subsection{Robot experiment and demonstration}
\label{sec:robotexperiment}
We implemented ROS nodes for transparent object depth completion using the depth-to-depth network and a Detectron2~\cite{wu2019detectron2} instance segmentation network fine-tuned on transparent object data generated using NeuralLabeling.
The robot that we used is RT Corporation Sciurus17.
Grasps are evaluated on two objects, a tall glass and a wine glass, which were part of the dataset for training the depth completion network and fine-tuning the segmentation network.
We placed the objects in 9 positions inside the dishwasher, and performed 3 trials per position, for a total of 54 trials.
The overall grasp success rate using the system is 83.3\%.
Wine glass grasp success rate was 92.3\% and tall glass grasp success rate was 75\%.
We performed the same experiment with our prediction segmentation masks but without using depth completion (i.e.\ the original sensor depth).
The overall grasp success rate without depth completion was 16.3\%.
Wine glass grasp success rate without depth completion was 29.6\% and tall glass grasp success rate without depth completion was 0\%.
In future work, we plan to explore more advanced neural network designs for more accurate depth completion, as well as mechanical improvements to the gripper to allow for a larger error tolerance.
As shown in the supplemental material, our robot system can also perform sequential grasping of transparent objects placed in a dishwasher environment.

\section{DISCUSSION AND CONCLUSION}
We presented \emph{NeuralLabeling}, a labeling approach and toolset for annotating NeRF renderings and generating datasets for downstream deep learning applications.
With \emph{NeuralLabeling} we were able to rapidly create datasets of transparent objects in a complex environment and use the datasets to greatly improve the performance of transparent object depth completion and to perform instance segmentation in a transparent object manipulation example.
The main limitation of \emph{NeuralLabeling} is the significant time required to record scenes and generate camera extrinsics for each captured frame, however this is mostly automated and could be further automated in the future.
In future we plan to apply \emph{NeuralLabeling} to larger scenes such as supermarkets and convenience stores for generating datasets to fine-tune vision-language models.
We also plan to investigate how \emph{NeuralLabeling} can be applied to dynamic scenes and how high-quality object meshes can be used to insert objects into scenes where the objects were not originally located.


\section*{ACKNOWLEDGMENT}

The authors would like to thank Lukas Meyer for the fruitful discussions.
This work was supported by JST [Moonshot R\&D][Grant Number JPMJMS2031].
This research is subsidized by New Energy and Industrial Technology Development Organization (NEDO) under a project JPNP20016.
This paper is one of the achievements of joint research with and is jointly owned copyrighted material of ROBOT Industrial Basic Technology Collaborative Innovation Partnership.

\printbibliography

\end{document}